%% file: main.tex
\documentclass[runningheads]{llncs}
\usepackage{amsmath,amssymb} 
\usepackage{color}
\usepackage[width=122mm,left=12mm,paperwidth=146mm,height=193mm,top=12mm,paperheight=217mm]{geometry}

\usepackage[utf8]{inputenc} 
\usepackage[T1]{fontenc}    
\usepackage{booktabs}       
\usepackage{amsfonts}       
\usepackage{nicefrac}       
\usepackage{microtype}      
\usepackage{color}
\usepackage{colortbl}
\usepackage{tabularx}
\usepackage{subcaption}
\usepackage{arydshln}
\usepackage{array,multirow}
\usepackage{xfrac}
\usepackage{bm}
\usepackage{soul}
\usepackage[export]{adjustbox}
\usepackage{graphicx}
\usepackage{verbatim}
\usepackage{mwe}
\usepackage[shortcuts]{extdash}
\usepackage[toc,page]{appendix}
\usepackage[ruled,lined,linesnumberedhidden]{algorithm2e}
\usepackage{enumitem}
\usepackage{bbold}
\usepackage[normalem]{ulem}
\usepackage[dvipsnames]{xcolor}

\DeclareMathOperator*{\argmax}{argmax} 
\newcommand\ddfrac[2]{\frac{\displaystyle #1}{\displaystyle #2}}

\newcolumntype{C}[1]{>{\centering\let\newline\\\arraybackslash\hspace{0pt}}m{#1}}
\newcolumntype{K}[1]{>{\centering\arraybackslash}p{#1}}

\newcommand{\hsseo}[1]{\textcolor{blue}{#1}}

\newcommand{\ie}{\textit{i}.\textit{e}.}

\newcommand{\beginsupplement}{%
        \setcounter{table}{0}
        \renewcommand{\thetable}{\Alph{table}}%
        \setcounter{figure}{0}
        \renewcommand{\thefigure}{\Alph{figure}}%
        \setcounter{section}{0}
        \renewcommand{\thesection}{\Alph{section}}%
     }

\begin{document}

\pagestyle{headings}
\mainmatter
\def\ECCV18SubNumber{2333}  

\title{CPlaNet: Enhancing Image Geolocalization \\ by Combinatorial Partitioning of Maps} 

\titlerunning{Enhancing Image Geolocalization by Combinatorial Partitioning of Maps}

\authorrunning{P. H. Seo, T. Weyand, J. Sim, and B. Han}

\author{Paul Hongsuck Seo\inst{1} \and Tobias Weyand\inst{2} \and Jack Sim\inst{2} \and Bohyung Han\inst{3}}
\institute{Dept. of CSE, POSTECH, Korea \and
           Google Research, USA \and 
           Dept. of ECE \& ASRI, Seoul National University, Korea \\
           \email{hsseo@postech.ac.kr}
           \email{  \{weyand,jacksim\}@google.com}
           \email{  bhhan@snu.ac.kr}
           }

\maketitle

\input{sections/abstract.tex}

\input{sections/introduction.tex}
\input{sections/related_work.tex}
\input{sections/comb_part.tex}

\input{sections/learning.tex}

\input{sections/experiments.tex}

\input{sections/conclusion.tex}

\section*{Acknowledgment}
The part of this work was performed while the first and last authors were with Google, Venice, CA.
This research is partly supported by the IITP grant [2017-0-01778], and the Technology Innovation Program [10073166] funded by the Korea government MSIT and MOTIE, respectively.

\bibliographystyle{splncs}
\bibliography{egbib}

\clearpage

\beginsupplement
\section*{\em Supplementary Material}
\begin{appendix}

\section{Inference Complexity}
We show the additional complexity for the combinatorial partitioning is theoretically negligible in Section 5.4.2 of the main document
To support the analysis, we measure the inference time of our model with five base classifiers and compare it with the result from a single classifier model.
Table~\ref{tab:time_complexity} presents that time for feature extraction is the most dominant factor in inference time while classification layers have minor overhead in prediction.

\begin{table}[h!]
\centering
\caption{
Average inference time of a single classifier and five classifiers.
}
\label{tab:time_complexity}
\begin{tabular}{
@{}C{0.2cm}@{}
@{}p{4cm}@{}|@{}C{3cm}@{}
}
&                           & inference time (sec)  \\ \hline
&feature extraction (FE)    & 0.11                  \\
&FE+1 classifier            & 0.12                  \\
&FE+5 classifiers           & 0.13                 \\
\hline
\end{tabular}
\end{table}

\begin{figure}[t]
    \centering
    \begin{subfigure}[b]{0.3\linewidth}
        \centering
        \includegraphics[width=1\linewidth] {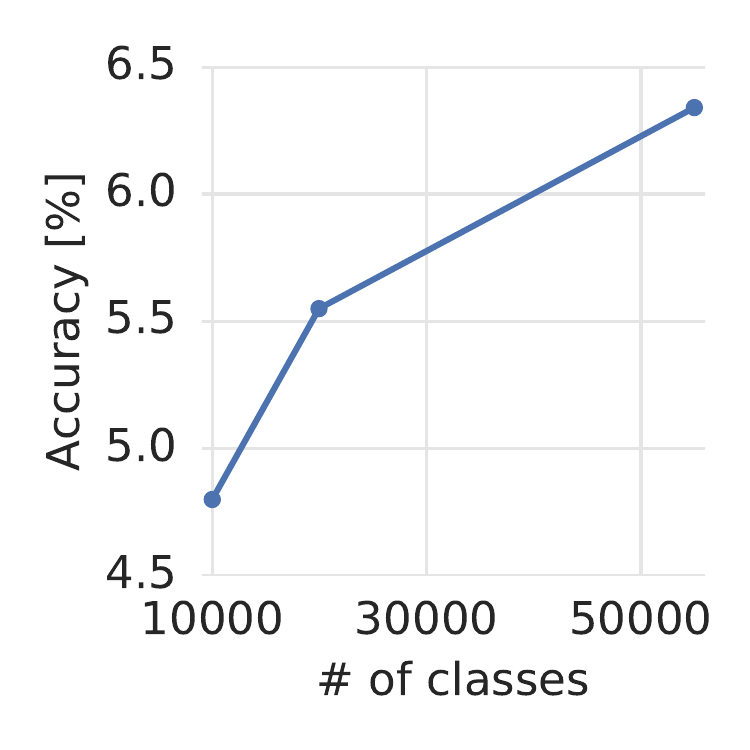}
        \caption{1~km}
    \end{subfigure}
    \begin{subfigure}[b]{0.3\linewidth}
        \centering
        \includegraphics[width=1\linewidth] {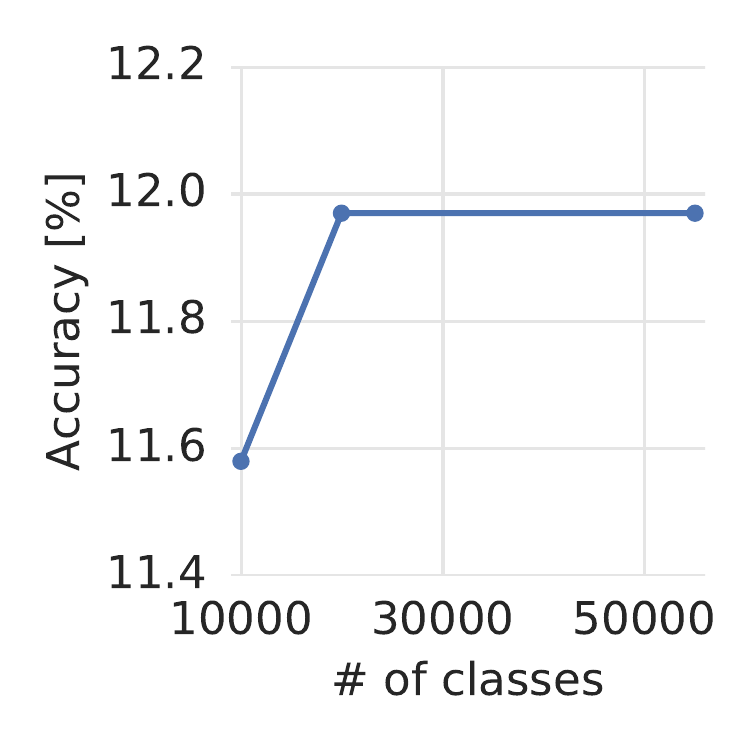}
        \caption{25~km}
    \end{subfigure}
    \begin{subfigure}[b]{0.3\linewidth}
        \centering
        \includegraphics[width=1\linewidth] {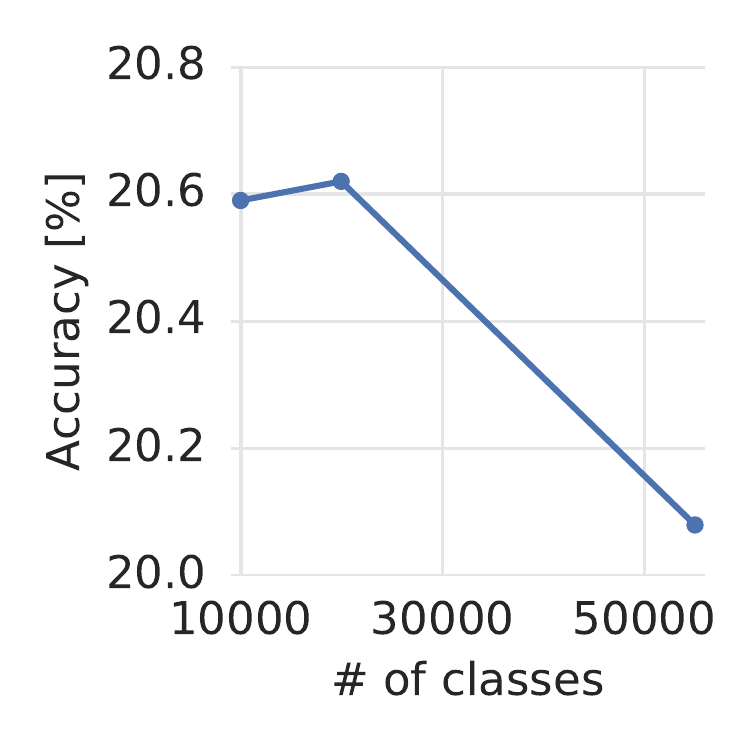}
        \caption{200~km}
    \end{subfigure}
    \begin{subfigure}[b]{0.3\linewidth}
        \centering
        \includegraphics[width=1\linewidth] {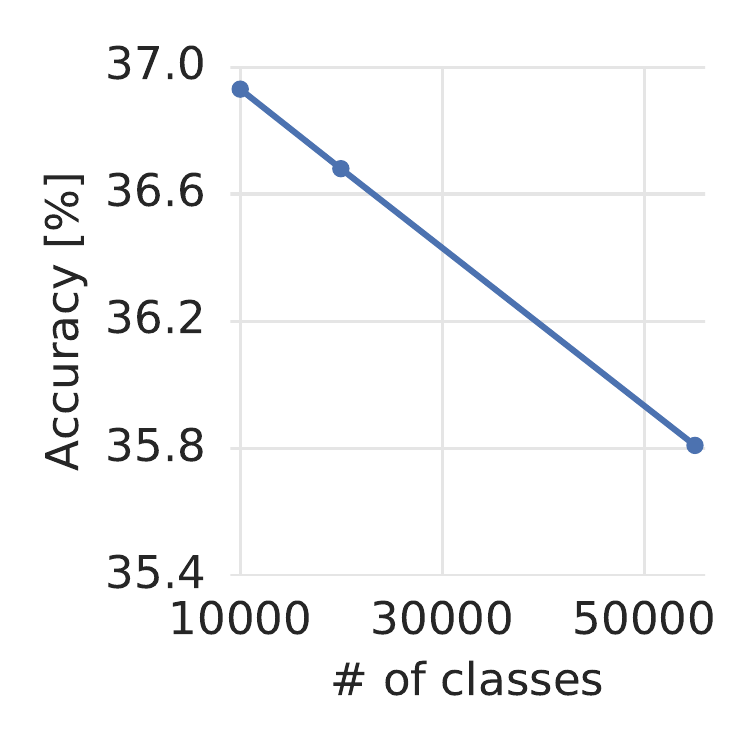}
        \caption{750~km}
    \end{subfigure}
    \begin{subfigure}[b]{0.3\linewidth}
        \centering
        \includegraphics[width=1\linewidth] {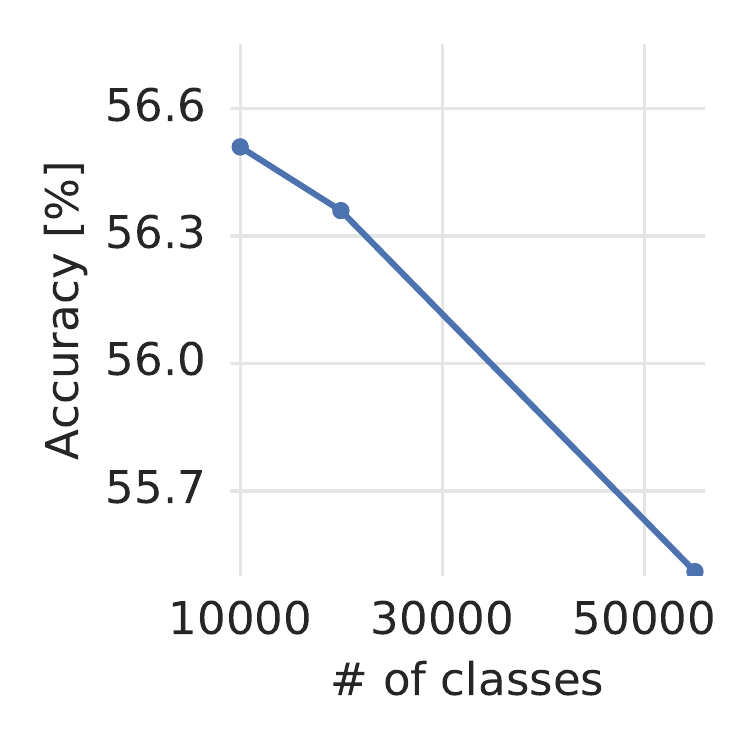}
        \caption{2500~km}
    \end{subfigure}
    \caption{
    Geolocalization accuracies of a single-head classifier at five different distance thresholds---1~km, 25~km, 200~km, 750~km, and 2500~km---by varying the number of geoclasses of the classifier. 
    The accuracy with 1~km increases as more geoclasses are employed, but the benefit of using more classes is gradually reduced and even becomes negative as the thresholds increase.
    Note that the geoclasses are generated by our generation algorithm using the parameters for ClassSet~1 in the main paper except for the number of classes.
    }
    \label{fig:acc_graph}
\end{figure}

\section{Number of Geoclasses and Data Deficiency}
In classification-based image geolocalization, the number of geoclasses in the classifier is closely related to precision of a prediction.
In other words, since the geolocation of an image is often given by the center of its predicted geoclass, a coarse-grained partitioning inherently have large errors quantitatively due to its low resolution.
So, it is preferable to have more geoclasses by a fine-grained partitioning.
However, it is not always straightforward to increase the number of classes due to training data deficiency; as the partitions are more fine-grained, the number of training examples per geoclass decreases.

Figure~\ref{fig:acc_graph} presents image geolocalization accuracy at five different levels of distance thresholds while varying the number of geoclasses in a classifier. 
According to our experiment, the accuracy with a small distance threshold typically improves when trained with more geoclasses whereas the accuracy with a large distance threshold decreases.
We believe that such inconsistent phenomenon results from skewed distribution of image geolocations over the map.
Since images in geoclasses with dense image population often contain common landmarks and share visual features,
dividing these geoclasses into more fine-grained ones leads to reducing the prediction error.
On the other hand, images are more heterogeneous in sparse geoclasses and subdividing these geoclasses leads to the data deficiency problem causing accuracy drops.
Note that, since predictions on sparse geoclasses are unlikely to be very accurate in coarse-grained partitioning, further subdivisions do no harm to the the low-threshold results and accuracy drops mostly happen in high-threshold areas.

Thus, it is not always desirable to simply increase the number of geoclasses for improving performance.
In contrast, our method achieves the highest geolocalization accuracy at almost every threshold level with an increased number of distinct geoclasses given by combinatorial partitioning.
Note that combinatorial partitioning enables the model to work around the data deficiency problem.
We also emphasize that we can apply our method to any base classifiers even with different design choices.

\begin{figure}[t]
    \centering
    \begin{subfigure}[m]{0.45\linewidth}
    \includegraphics[width=\linewidth]{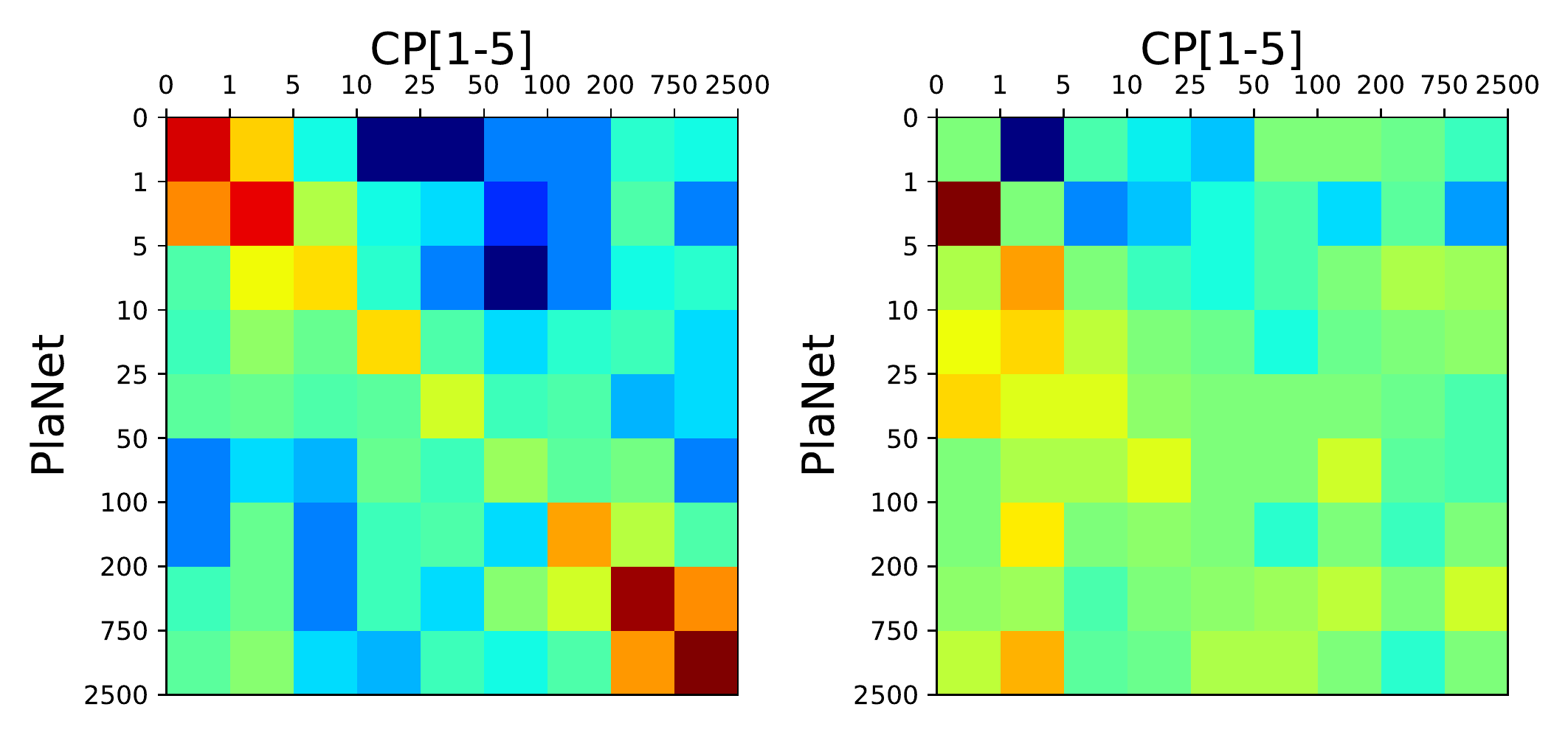}
    \end{subfigure}~~~~
    \begin{subfigure}[m]{0.42\linewidth}
    \includegraphics[height=2cm]{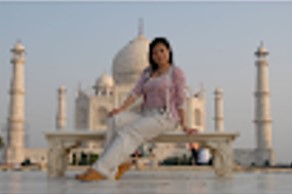}
    \includegraphics[height=2cm]{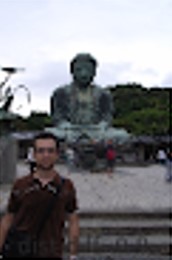}
    \end{subfigure}
    \vspace{-0.4cm}
    \caption{ \small
    Qualitative comparison of our algorithm and PlaNet. (left) 2D heat map of prediction quality on Im2GPS3k.
    Axis ticks denote geo-distance bins.
    (right) images from bin (1,5] of PlaNet to bin (0,1] of CP[1-5].
    }
    \label{fig:heatmap}
    \vspace{-0.6cm}
\end{figure}

\vspace{-0.5cm}
\section{Qualitative Evaluation}
We conducted qualitative analysis comparing CP[1-5] and PlaNet~(reprod).
Figure~\ref{fig:heatmap}(left) presents a 2D matrix ($A$) made by counting the number of geolocalization prediction pairs corresponding to each element given by the two models on Im2GPS3k dataset while Figure~\ref{fig:heatmap}(middle) shows another matrix ($A-A^\text{T}$) whose lower triangle shows how much CP[1-5] improves accuracy with respect to PlaNet.
According to our observation, the gain is most significant in the pair of bin corresponding to (1,5] of PlaNet and bin corresponding to (0,1] of CP[1-5].
The images that belong to the observation frequently contain landmark photos as shown in Figure~\ref{fig:heatmap}(right).

\end{appendix}

\end{document}

%% file: sections/abstract.tex

\begin{abstract}
Image geolocalization is the task of identifying the location depicted in a photo based only on its visual information. 
This task is inherently challenging since many photos have only few, possibly ambiguous cues to their geolocation.
Recent work has cast this task as a classification problem by partitioning the earth into a set of discrete cells that correspond to geographic regions.
The granularity of this partitioning presents a critical trade-off; using fewer but larger cells results in lower location accuracy while using more but smaller cells reduces the number of training examples per class and increases model size, making the model prone to overfitting.
To tackle this issue, we propose a simple but effective algorithm, combinatorial partitioning, which generates a large number of fine-grained output classes by intersecting multiple coarse-grained partitionings of the earth.
Each classifier votes for the fine-grained classes that overlap with their respective coarse-grained ones. 
This technique allows us to predict locations at a fine scale while maintaining sufficient training examples per class.
Our algorithm achieves the state-of-the-art performance in location recognition on multiple benchmark datasets.
\keywords{Image geolocalization, combinatorial partitioning, fine-grained classification}
\end{abstract}

%% file: sections/introduction.tex

\section{Introduction}
\label{sec:introduction}

Image geolocalization is the task of predicting the geographic location of an image based only on its pixels without any meta-information.
As the geolocation is an important attribute of an image by itself, it also plays as a proxy to other location attributes such as elevation, weather, and distance to a particular point of interest.
However, geolocalizing images is a challenging task since input images often contain limited visual information representative of their locations.
To handle this issue effectively, the model is required to capture and maintain visual cues of the globe comprehensively.

There exist two main streams to address this task: retrieval and classification based approaches.
The former searches for nearest neighbors in a database of geotagged images by matching their feature representations~\cite{hays08im2gps,hays15large,vo17revisiting}.
Visual appearance of an image at a certain geolocation is estimated using the representations of the geotagged images in database.
The latter treats the task as a classification problem by dividing the map into multiple discrete classes~\cite{vo17revisiting,weyand16planet}.
Thanks to recent advances in deep learning, simple classification techniques based on convolutional neural networks handle such complex visual understanding problems effectively.

There are several advantages of formulating the task as classification instead of retrieval.
First, classification-based approaches save memory and disk space to store information for geolocalization; they just need to store a set of model parameters learned from training images whereas all geotagged images in the database should be embedded and indexed to build retrieval-based systems.
In addition to space complexity, inference of classification-based approaches is faster because a result is given by a simple forward pass computation of a deep neural network while retrieval-based methods undergo significant overhead for online search from a large index given a query image.
Finally, classification-based algorithms provide multiple hypotheses of geolocation with no additional cost by presenting multi-modal answer distributions.

On the other hand, the standard classification-based approaches have a few critical limitations.
They typically ignore correlation of spatially adjacent or proximate classes.
For instance, assigning a photo of Bronx to Queens, which are both within New York city, is treated equally wrong as assigning it to Seoul.
Another drawback comes from artificially converting continuous geographic space into discrete class representations.
Such an attempt may incur various artifacts since images near class boundaries are not discriminative enough compared to data variations within classes; training converges slowly and 
performance is affected substantially by subtle changes in map partitioning.
This limitation can be alleviated by increasing the number of classes and reducing the area of the region corresponding to each class.
However, this strategy increases the number of parameters while decreasing the size of the training dataset per class.

\begin{figure}[t]
\centering
\includegraphics[width=0.8\linewidth]{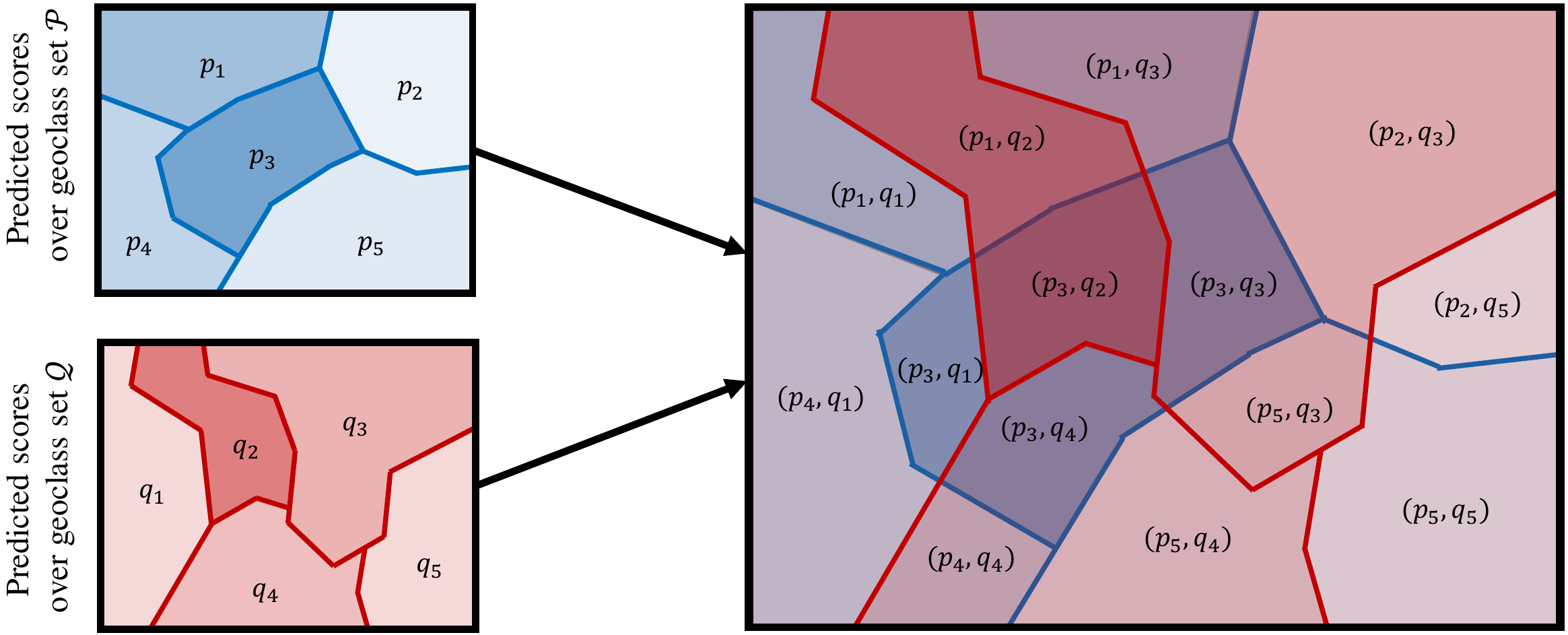}
\caption{
Visualization of combinatorial partitioning. 
Two coarse-grained class sets, $\mathcal{P}=\{p_1,p_2,\dots,p_5\}$ and $\mathcal{Q}=\{q_1,q_2,\dots,q_5\}$ in the map on the left, are merged to construct a fine-grained partition as shown in the map on the right by a combination of geoclasses in the two class sets.
Each resulting fine-grained class is represented by a tuple $(p_i, q_j)$, and is constructed by identifying partially overlapping partitions in $\mathcal{P}$ and $\mathcal{Q}$.
}
\label{fig:comb_part}
\end{figure}

To overcome such limitations, we propose a novel algorithm that enhances the resolution of geoclasses and avoids the training data deficiency issue.
This is achieved by \emph{combinatorial partitioning}, which is a simple technique to generate spatially fine-grained classes through a combination of the multiple configurations of classes.
This idea has analogy to product quantization~\cite{jegou11product} since they both construct a lot of quantized regions using relatively few model parameters through a combination of low-bit subspace encodings or coarse spatial quantizations.
Our combinatorial partitioning allows the model to be trained with more data per class by considering a relatively small number of classes at a time.
Figure~\ref{fig:comb_part} illustrates an example of combinatorial partitioning, which enables generating more classes with minimal increase of model size and learning individual classifiers reliably without losing training data per class.
Combinatorial partitioning is applied to an existing classification-based image geolocalization technique, PlaNet~\cite{weyand16planet}, and our algorithm is referred to as CPlaNet hereafter.
%
Our contribution is threefold:
\begin{itemize}
    \item[$\bullet$] We introduce a novel classification-based model for image geolocalization using combinatorial partitioning, which defines a fine-grained class configuration by combining multiple heterogeneous geoclass sets in coarse levels.
    \item[$\bullet$] We propose a technique that generates multiple geoclass sets by varying parameters, and design an efficient inference technique to combine prediction results from multiple classifiers with proper normalization.
    \item[$\bullet$] The proposed algorithm outperforms the existing techniques in multiple benchmark datasets, especially at fine scales.
\end{itemize}

The rest of this paper is organized as follows. 
We review the related work in Section~\ref{sec:rel_work}, and describe combinatorial partitioning for image geolocalization in Section~\ref{sec:comb_part}.
The details about training and inference procedures are discussed in Section~\ref{sec:learning}.
We present experimental results of our algorithm in Section~\ref{sec:exp}, and conclude our work in Section~\ref{sec:conclusion}.

%% file: sections/related_work.tex

\section{Related Work}
\label{sec:rel_work}

%
%
%
%
%
%
The most common approach of image geolocalization is based on the image retrieval pipeline. 
Im2GPS \cite{hays08im2gps,hays15large} and its derivative~\cite{vo17revisiting} perform image retrieval in a database of geotagged images using global image descriptors. 
%
Various visual features can be applied to the image retrieval step.
NetVLAD~\cite{Arandjelovic16CVPR} is a global image descriptor trained end-to-end for place recognition on street view data using a ranking loss. 
Kim et al.~\cite{Kim17CVPR} learn a weighting mask for the NetVLAD descriptor to focus on image regions containing location cues. 
While global features have the benefit to retrieve diverse natural scene images based on ambient information, local image features yield higher precision in retrieving structured objects such as buildings and are thus more frequently used~\cite{Baatz10ECCV,Cao15IJCV,Chen11CVPR,Kim15ICCV,Knopp10ECCV,Philbin07CVPR,Schindler07CVPR,Zamir10ECCV,Zamir14PAMI}. 
DELF~\cite{Noh17ICCV} is a deeply learned local image feature detector and descriptor with attention for image retrieval.

On the other hand, classification-based image geolocalization formulates the problem as a classification task. 
In~\cite{vo17revisiting,weyand16planet}, a classifier is trained to predict the geolocation of an input image.
Since the geolocation is represented in a continuous space, classification-based approaches quantize the map of the entire earth into a set of geoclasses corresponding to partitioned regions.
Note that training images are labeled into the corresponding geoclasses based on their GPS tags.
At test time, the center of the geoclass with the highest score is returned as the predicted geolocation of an input image.
This method is lightweight in terms of space and time complexity compared to retrieval-based methods, but its prediction accuracy highly depends on how the geoclass set is generated.
Since every image that belongs to the same geoclass has an identical predicted geolocation, more fine-grained partitioning is preferable to obtain precise predictions.
However, it is not always straightforward to increase the number of geoclasses as it linearly increases the number of parameters and makes the network prone to overfitting to training data.

%
%
Pose estimation approaches~\cite{Cao15IJCV,Irschara09CVPR,Li10ECCV,Li12ECCV,Liu17ICCV,Sattler11ICCV,Sattler12BMVC} match  query images against 3D models of an area, and employ 2D-3D feature correspondences to identify 6-DOF query poses. 
Instead of directly matching against a 3D model, \cite{Sattler17CVPR,Sattler12BMVC} first perform image retrieval to obtain coarse locations and then estimate poses using the retrieved images.
%
%
%
PoseNet~\cite{Kendall17CVPR,Kendall15ICCV} treats pose estimation as a regression problem based on a convolutional neural network.
The accuracy of PoseNet is improved by introducing an intermediate LSTM layer for dimensionality reduction~\cite{Walch17ICCV}.

%
%
A related line of research is landmark recognition, where images are clustered by their geolocations and visual similarity to construct a database of popular landmarks.
The database serves as the index of an image retrieval system~\cite{Avrithis10MM,Gammeter09ICCV,Johns11ICCV,Quack08CIVR,Zheng09CVPR,Weyand15CVIU} or the training data of a landmark classifier~\cite{Bergamo13CVPR,Li09CVPR,Gronat13CVPR}.
%
%
%
Cross-view geolocation recognition makes additional use of satellite or aerial imagery to determine query locations~\cite{Workman15ICCV,Lin13CVPR,Lin15CVPR,Tian17CVPR}.

%% file: sections/comb_part.tex

\section{Geolocalization using Multiple Classifiers}
\label{sec:comb_part}

Unlike existing classification-based methods~\cite{weyand16planet}, CPlaNet relies on multiple classifiers that are all trained with unique geoclass sets.
The proposed model predicts more fine-grained geoclasses, which are given by combinatorial partitioning of multiple geoclass sets.
Since our method requires a distinct geoclass set for each classifier, we also propose a way to generate multiple geoclass sets.

\subsection{Combinatorial Partitioning}
\label{sec:sub:part}
Our primary goal is to establish fine-grained geoclasses through a combination of multiple coarse geoclass sets and exploit benefits from both coarse- and fine-grained geolocalization-by-classification approaches.
In our model, there are multiple unique geoclass sets represented by partitions $\mathcal{P}=\{p_1,p_2,\dots,p_5\}$ and $\mathcal{Q}=\{q_1,q_2,\dots,q_5\}$ as illustrated on the left side of Figure~\ref{fig:comb_part}.
Since the region boundaries in these geoclass sets are unique, overlapping the two maps constructs a set of fine-grained subregions.
This procedure, referred to as combinatorial partitioning, is identical to the Cartesian product of the two sets, but disregards the tuples given by two spatially disjoint regions in the map.
For instance, combining two aforementioned geoclass sets in Figure~\ref{fig:comb_part}, we obtain fine-grained partitions defined by a tuple $(p_i, q_j)$ as depicted by the map on the right of the figure, while the tuples made by two disjoint regions, {\it e.g.,} $(p_1, q_5)$, are not considered.

While combinatorial partitioning aggregates results from multiple classifiers, it is conceptually different from ensemble models whose base classifiers predict labels in the same output space.
In combinatorial partitioning, while each coarse-grained partition is scored by a corresponding classifier, fine-grained partitions are given different scores by the combinations of multiple unique geoclass sets.
Also, combinatorial partitioning is closely related to product quantization~\cite{jegou11product} for approximate nearest neighbor search in the sense that they both generate a large number of quantized regions by either a Cartesian product of quantized subspaces or a combination of coarse space quantizations. 
Note that combinatorial partitioning is a general framework and applicable to other tasks, especially where labels have to be defined on the same embedded space as in geographical maps.

\subsection{Benefits of Combinatorial Partitioning}
\label{sec:sub:benefits}
The proposed classification model with combinatorial partitioning has the following three major benefits.

\subsubsection{Fine-grained classes with fewer parameters}
Combinatorial partitioning generates fine-grained geoclasses using a smaller number of parameters because a single geoclass in a class set can be divided into many subregions by intersections with geoclasses from multiple geoclass sets.
For instance in Figure~\ref{fig:comb_part}, two sets with 5 geoclasses form 14 distinct classes by the combinatorial partitioning.
If we design a single flat classifier with respect to the fine-grained classes, it requires more parameters, \ie, $14 \times F > 2\times(5\times F)$, where $F$ is the number of input dimensions to the classification layers.

\subsubsection{More training data per class}
Training with fine-grained geoclass sets is more desirable for higher resolution of output space, but is not straightforward due to training data deficiency; the more we divide the maps, the less training images remain per geoclass.
Our combinatorial partitioning technique enables us to learn models with coarsely divided geoclasses and maintain more training data in each class than a na\"ive classifier with the same number of classes.


\subsubsection{More reasonable class sets}
There is no standard method to define geoclasses for image geolocalization, so that images associated with the same classes have common characteristics.
An arbitrary choice of partitioning may incur undesirable artifacts due to heterogeneous images located near class territories; the features trained on loosely defined class sets tend to be insufficiently discriminative and less representative.
On the other hand, our framework constructs diverse partitions based on various criteria observed in the images. 
We can define more tightly-coupled classes through combinatorial partitioning by distilling noisy information from multiple sources.

\begin{figure}[t]
\centering
\includegraphics[width=0.32\linewidth] {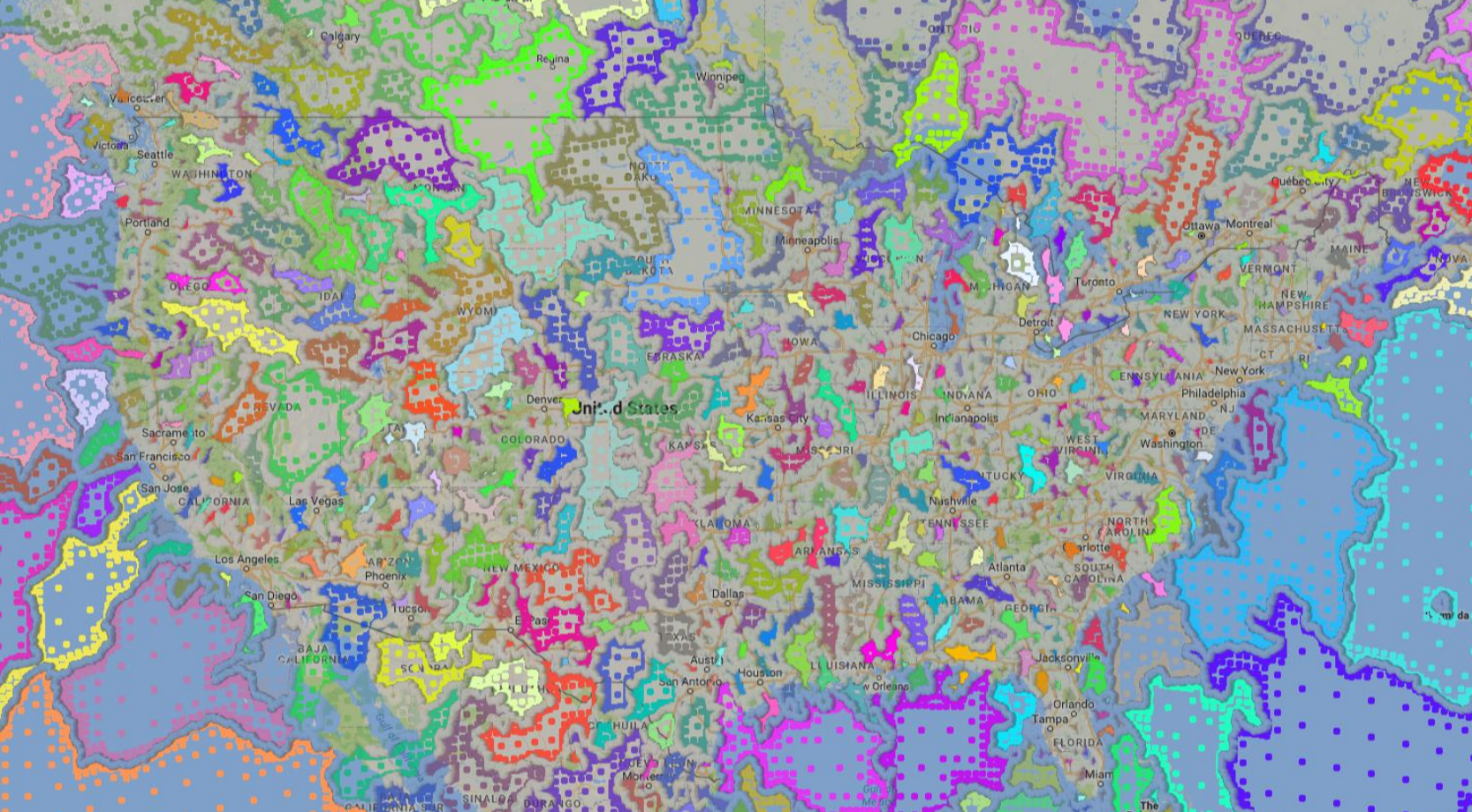}
\includegraphics[width=0.32\linewidth] {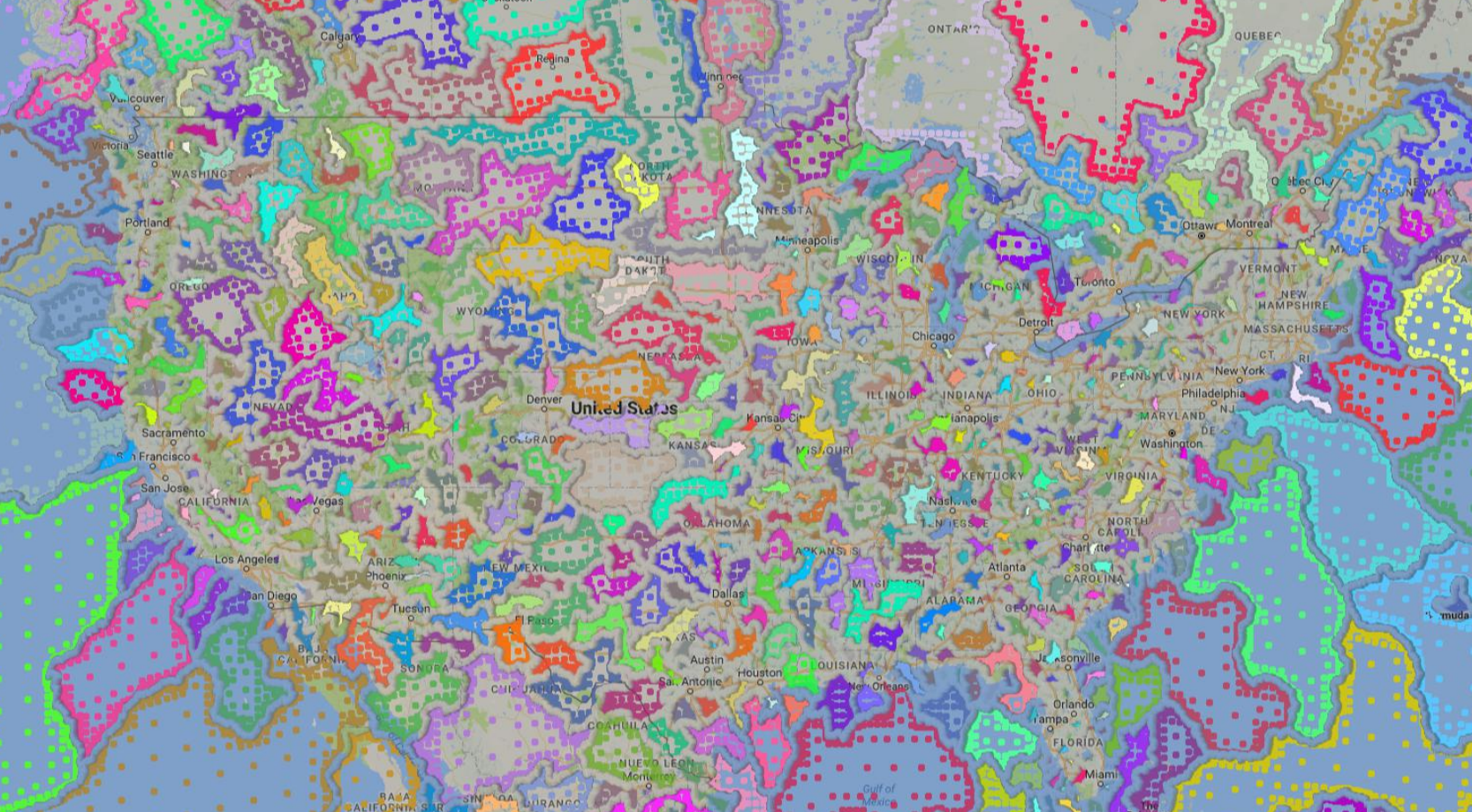} \\
\centerline{(a) Geoclass set 1 \quad\quad\quad\quad (b) Geoclass set 2}
\includegraphics[width=0.32\linewidth] {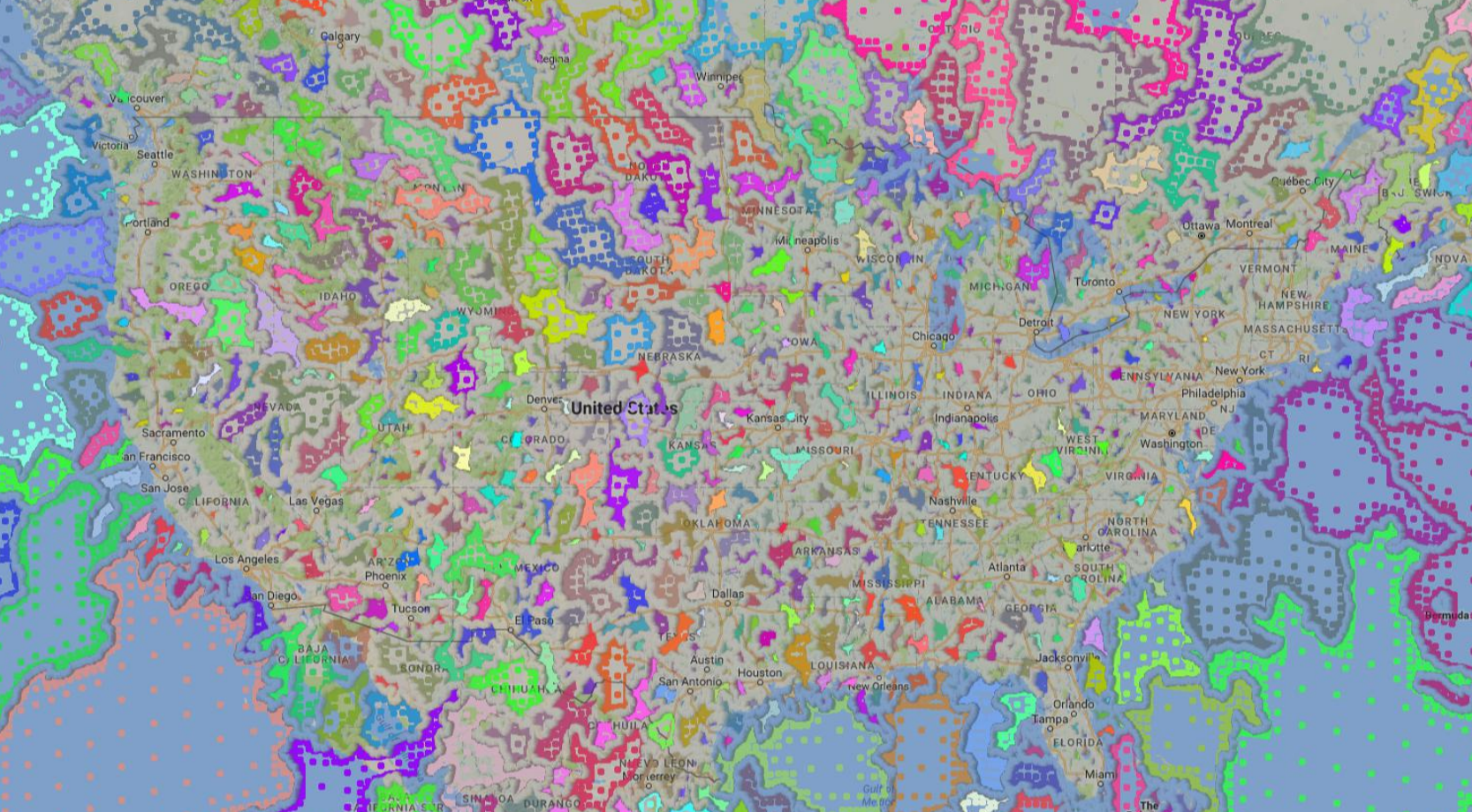}
\includegraphics[width=0.32\linewidth] {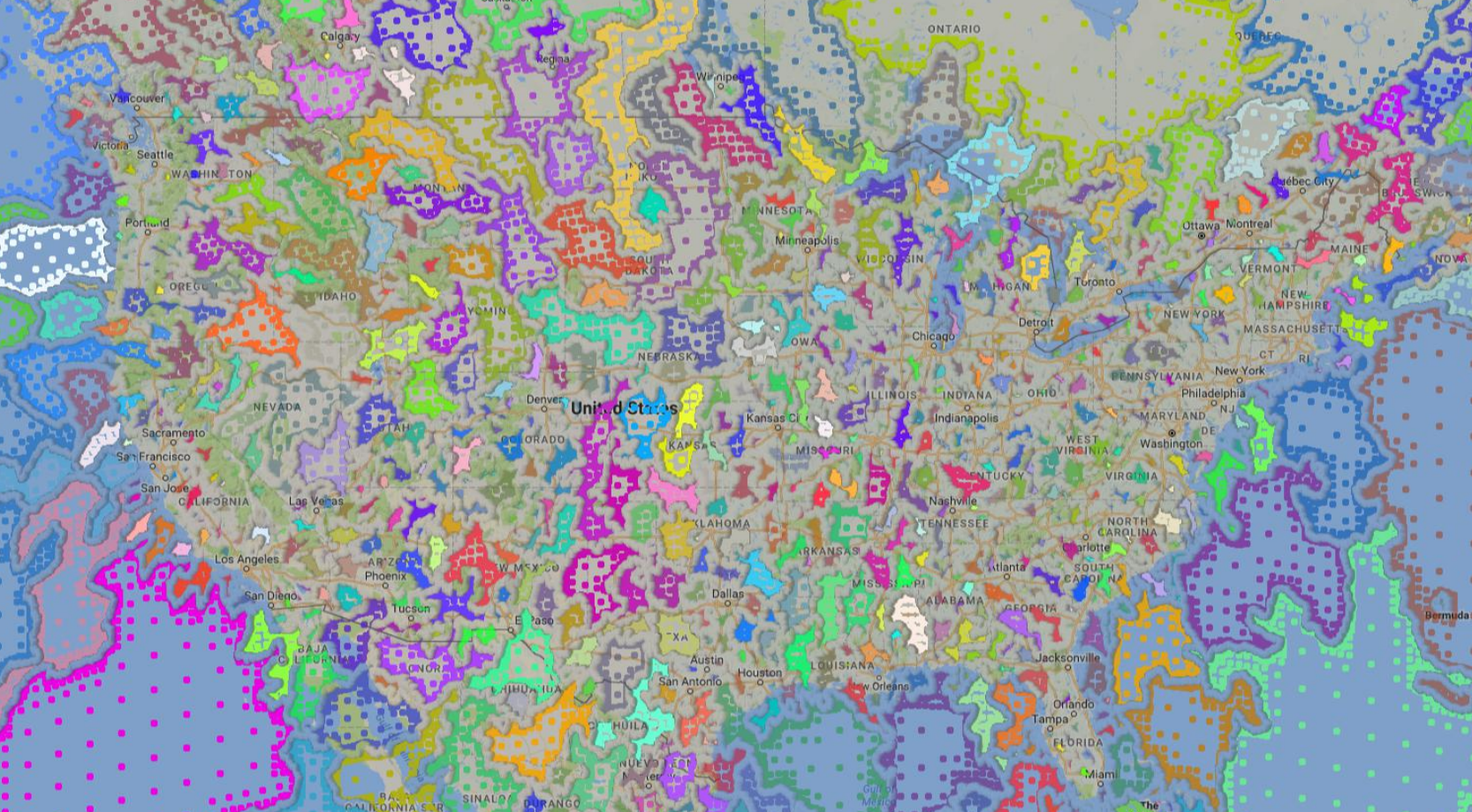}
\includegraphics[width=0.32\linewidth] {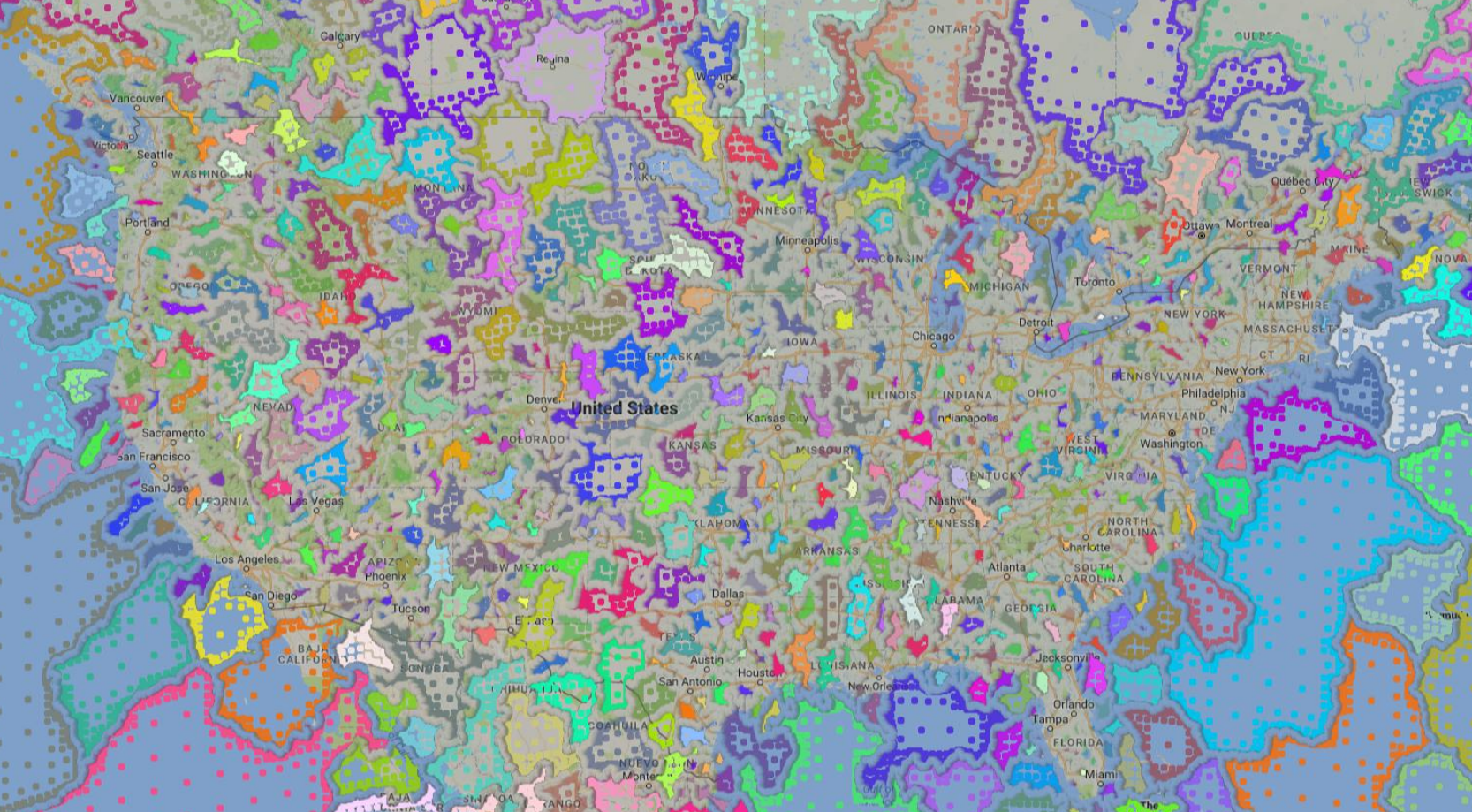}
\centerline{(c) Geoclass set 3 \quad\quad\quad\quad~ (d) Geoclass set 2 ~\quad\quad\quad\quad (3) Geoclass set 3}
\caption{
Visualization of the geoclass sets on the maps of the United States generated by the parameters shown in Table~\ref{tab:geoclass}. 
Each distinct region is marked by a different color.
The first two sets, (a) and (b), are generated by manually designed parameters while parameters for the others are randomly sampled.
}
\label{fig:class_set}
\end{figure}

\subsection{Generating Multiple Geoclass Sets}
\label{sec:sub:geo_gen}
The geoclass set organization is an inherently ill-posed problem as there is no consensus about ideal region boundaries for image geolocalization.
Consequently, it is hard to define the optimal class configuration, which motivates the use of multiple random boundaries in our combinatorial partitioning.
We therefore introduce a mutable method of generating geoclass sets, which considers both visual and geographic distances between images.

The generation method starts with an initial graph for a map, where a node represents a region in the map and an edge connects two nodes of adjacent regions.
We construct the initial graph based on S2 cells\footnote{We use Google's S2 library. S2 cells are given by a geographical partitioning of the earth into a hierarchy. The surface of the earth is projected onto six faces of a cube. Each face of the cube is hierarchically subdivided and forms S2 cells in a quad-tree. Refer to \url{https://code.google.com/archive/p/s2-geometry-library/ for more details.}} at a certain level.
Empty S2 cells, which contain no training image, do not construct separate nodes and are randomly merged with one of their neighboring non-empty S2 cells.
This initial graph covers the entire surface of the earth.
Both nodes and edges are associated with numbers---scores for nodes and weights for edges.
We give a score to each node by a linear combination of three different factors: the number of images in the node and the number of empty and non-empty S2 cells.
An edge weight is computed by the weighted sum of geolocational and visual distances between two nodes.
The geolocational distance is given by the distance between the centers of two nodes while the visual distance is measured by cosine similarity based on the visual features of nodes, which are computed by averaging the associated image features extracted from the bottleneck layer of a pretrained CNN.
Formally, a node score $\omega(\cdot)$ and an edge weight $\nu(\cdot, \cdot)$ are defined respectively as
\begin{align}
    \omega(v_i) &= \alpha_1 \cdot n_\mathrm{img}(v_i) + \alpha_2 \cdot n_\mathrm{S2+}(v_i) + \alpha_3 \cdot n_\mathrm{S2}(v_i) \\
    \nu(v_i, v_j) &= \beta_1 \cdot \mathrm{dist_{vis}}(v_i, v_j) + \beta_2 \cdot \mathrm{dist_{geo}}(v_i, v_j)
\end{align}
where $n_\mathrm{img}(v)$, $n_\mathrm{S2+}(v)$ and $n_\mathrm{S2}(v)$ are functions that return the number of images, non-empty S2 cells and all S2 cells in a node $v$, respectively, and $\mathrm{dist_{vis}}(\cdot,\cdot)$ and $\mathrm{dist_{geo}}(\cdot, \cdot)$ are the visual geolocational distances between two nodes.
Note that the weights $(\alpha_1, \alpha_2, \alpha_3)$ and $(\beta_1, \beta_2)$ are free parameters in $[0, 1]$.

\begin{table}[t]
\centering
\caption{Parameters for geoclass set generation. 
Parameters for geoclass set 1 and 2 are manually given while the ones for rest geoclass sets are randomly sampled.}
\scalebox{0.9}{
\begin{tabular}{
@{}C{1.8cm}@{}|@{}C{6.1cm}@{}|@{}C{1.1cm}@{}@{}C{1.1cm}@{}|@{}C{1.1cm}@{}@{}C{1.1cm}@{}@{}C{1.1cm}@{}
}
\multicolumn{6}{l}{}\\
Parameter group                 & Parameters                                     & 1         & 2         & 3         & 4         & 5         \\ \hline\hline
\multirow{2}{*}{N/A}            & Num. of geoclasses                              & 9,969     & 9,969     & 12,977    & 12,333    & 11,262    \\ 
                                & Image feature dimensions                & 2,048     & 0         & 1,187     & 1,113     & 14,98     \\ \hline
\multirow{3}{*}{\parbox{1.8cm}{\centering Node score}}     & Weight for num. of images ($\alpha_1$)             & 1.000     & 1.000     & 0.501     & 0.953     & 0.713     \\ 
                                & Weight for num. of non-empty S2 cells ($\alpha_2$) & 0.000     & 0.000     & 0.490     & 0.044     & 0.287     \\ 
                                & Weight for num. of S2 cells ($\alpha_3$)          & 0.000     & 0.000     & 0.009     & 0.003     & 0.000     \\ \hline 
\multirow{2}{*}{\parbox{1.8cm}{\centering Edge weight}}    & Weight for visual distance ($\beta_1$)          & 1.000     & 0.000     & 0.421     & 0.628     & 0.057     \\ 
                                & Weight for geographical distance ($\beta_2$)    & 0.000     & 1.000     & 0.579     & 0.372     & 0.943     \\ \hline
\end{tabular}
}
\label{tab:geoclass}
\end{table}

After constructing the initial graph, we merge two nodes hierarchically in a greedy manner until the number of remaining nodes becomes the desired number of geoclasses.
To make each geoclass roughly balanced, we select the node with the lowest score first and merge it with its nearest neighbor in terms of edge weight.
A new node is created by the merge process and corresponds to the region given by the union of two merged regions.
The score of the new node is set to the sum of the scores of the two merged nodes.

The generated geoclass sets are diversified by the following free parameters: 1) the desired number of final geoclasses, 2) the weights of the factors in the node scores, 3) the weights of the two distances in computing edge weights and 4) the image feature extractor.
Each parameter setting constructs a unique geoclass set.
Note that multiple geoclass set generation is motivated by the fact that geoclasses are often ill-defined and the perturbation of class boundaries is a natural way to address the ill-posed problem.
Figure~\ref{fig:class_set} illustrates generated geoclass sets using different parameters described in Table~\ref{tab:geoclass}.

%% file: sections/learning.tex

\section{Learning and Inference}
\label{sec:learning}
This section describes more details about CPlaNet including network architecture, and training and testing procedure.
We also discuss data structures and the detailed inference algorithm. 

\subsection{Network Architecture}
\label{sub:training}
\begin{figure}[t]
\centering
\includegraphics[width=0.7\linewidth]{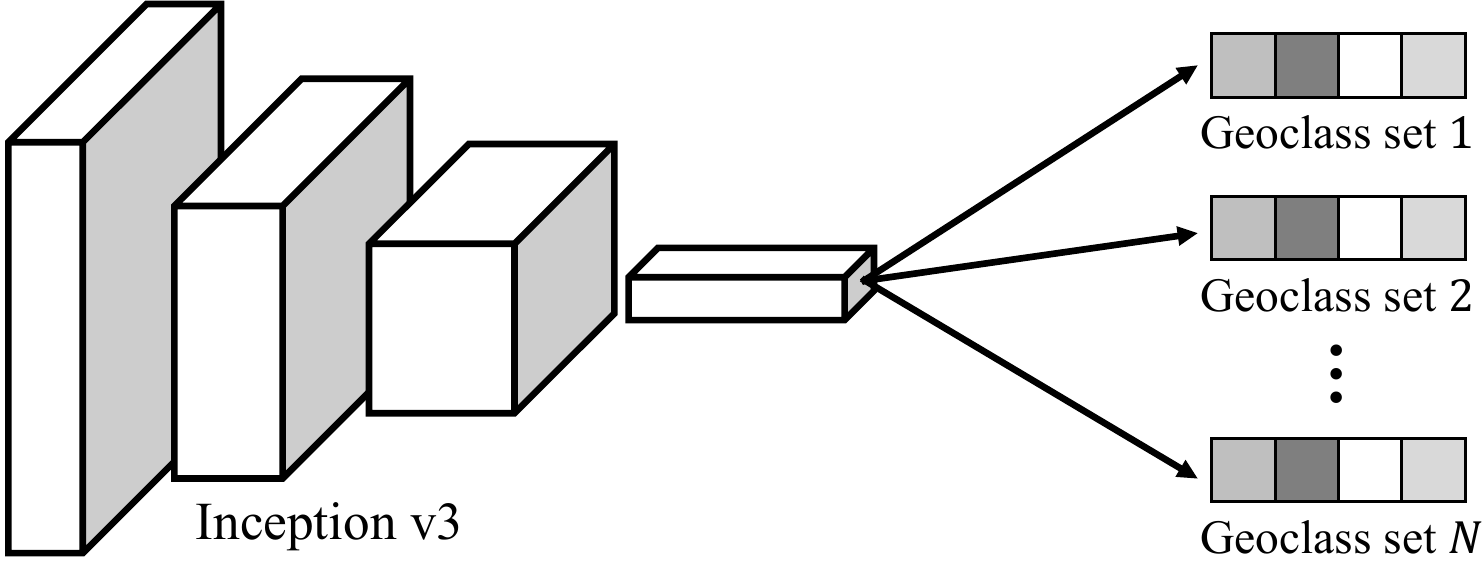}
\caption{
Network architecture of our model. 
A single Inception v3 architecture is used as our feature extractor after removing the final classification layer.
An image feature is fed to multiple classification branches and classification scores are predicted over multiple geoclass sets.
}
\label{fig:net_arch}
\end{figure}

Following \cite{weyand16planet}, we construct our network based on the Inception architecture~\cite{szegedy2015going} with batch normalization~\cite{szegedy2016rethinking}.
Inception v3 without the final classification layer (\texttt{fc} with \texttt{softmax}) is used as our feature extractor, and multiple branches of classification layers are attached on top of the feature extractor as illustrated in Figure~\ref{fig:net_arch}.
We train the multiple classifiers independently while keeping the weights of the Inception module fixed.
%
Note that, since all classifiers share the feature extractor, our model requires marginal increase of memory to maintain multiple classifiers.

\subsection{Inference with Multiple Classifiers}
\label{sub:inference}
Once the predicted scores in each class set are assigned to the corresponding regions, the subregions overlapped by multiple class sets are given cumulated scores from multiple classifiers.
%
A simple strategy to accumulate geoclass scores is to add the scores to individual S2 cells within the geoclass.
Such a simple strategy is inappropriate since it gives favor to classifiers that have geoclasses corresponding to large regions covering more S2 cells.
To make each classifier contribute equally to the final prediction regardless of its class configuration, we normalize the scores from individual classifiers with consideration of the number of S2 cells per class before adding them to the current S2 cell scores.
Formally, given a geoclass score distributed to S2 cell $g_k$ within a class in a geoclass set $\mathcal{C}^i$, denoted by $\mathrm{geoscore}(g_k; \mathcal{C}^i)$, an S2 cell is given a score $s(\cdot)$ by
\begin{equation}
    \label{eq:norm}
    s(g_k) = \sum_{i=1}^{N}{\frac{\mathrm{geoscore}(g_k; \mathcal{C}^i)}{\sum_{t=1}^{K}{\mathrm{geoscore}(g_t; \mathcal{C}^i)}}},
\end{equation}
where $K$ is the total number of S2 cells and $N$ is the number of geoclass sets.
Note that this process implicitly creates fine-grained partitions because the regions defined by different geoclass combinations are given different scores.

After this procedure, we select the S2 cells with the highest scores and compute their center for the final prediction of geolocation by averaging locations of images in the S2 cells. 
That is, the predicted geolocation $l_\mathrm{pred}$ is given by
%
%
\begin{equation}
    l_\mathrm{pred} = \frac{\sum_{k \in \mathcal{G}}{\sum_{e \in g_k}{\mathrm{geolocation}\left(e\right)}}}{\sum_{k \in \mathcal{G}}{\left| g_k \right|}},
    \label{eq:prediction}
\end{equation}
%
where $\mathcal{G} = \argmax_{k}{s(g_k)}$ is an index set of the S2 cells with the highest scores and $\mathrm{geolocation}(\cdot)$ is a function to return the ground-truth GPS coordinates of a training image $e$.
Note that an S2 cell $g_k$ may contain a number of training examples.

In our implementation, all fine-grained partitions are precomputed offline by generating all existing combinations of the multiple geoclass sets, and an index mapping from each geoclass to its corresponding partitions is also constructed offline to accelerate inference.
Moreover, we precompute the center of images in each partition.
To compute the center of a partition, we convert the latitude and longitude values of GPS tags into 3D Cartesian coordinates.
This is because a na\"ive average of latitude and longitude representations introduces significant errors as the target locations become distant from the equator.

%% file: sections/experiments.tex

\section{Experiments}
\label{sec:exp}

\subsection{Datasets}
\label{sub:datasets}
We train our network using a private dataset collected from Flickr, which has 30.3M geotagged images for training. 
We have sanitized the dataset by removing noisy examples to weed out unsuitable photos. 
For example, we disregard unnatural images (e.g., clipart images, product photos, etc.) and accept photos with a minimum size of 0.1 megapixels.
%

For evaluation, we mainly employ two public benchmark datasets---Im2GPS3k and YFCC4k~\cite{vo17revisiting}. 
The former contains 3,000 images from the Im2GPS dataset whereas the latter has 4,000 random images from the YFCC100m dataset.
In addition, we also evaluate on Im2GPS test set~\cite{hays08im2gps} to compare with previous work.
Note that Im2GPS3k is a different test benchmark from the Im2GPS test set.

\subsection{Parameters and Training Networks}
\label{sub:parameters}
We generate three geoclass sets using randomly generated parameters, which are summarized in Table~\ref{tab:geoclass}.
The number of geoclasses for each set is approximately between 10K and 13K, and the generation parameters for edge weights and node scores are randomly sampled.
Specifically, we select random axis-aligned subspaces out of the full 2,048 dimensions for image representations to diversify dissimilarity metrics between image representations.
Note that the image representations are extracted by a reproduced PlaNet~\cite{weyand16planet} after removing the final classification layer.
In addition to these geoclass sets, we generate two more sets with manually designed parameters; the edge weights in these two cases are given by either visual or geolocational distance exclusively, and their node scores are based on the number of images to mimic the setting of PlaNet.
Figure~\ref{fig:class_set} visualizes five geoclass sets generated by the parameters presented in Table~\ref{tab:geoclass}.

We use S2 cells at level 14 to construct the initial graph, where a total of $\sim$2.8M nodes are obtained after merging empty cells to their non-empty neighbors.
%
To train the proposed model, we employ the pretrained model of the reproduced PlaNet with its parameters fixed while the multiple classification branches are randomly initialized and fine-tuned using our training dataset.
The network is trained by RMSprop with a learning rate of 0.005.

\subsection{Evaluation Metrics}
Following \cite{vo17revisiting,weyand16planet}, we evaluate the models using geolocational accuracies at multiple scales by varying the allowed errors in terms of distances from ground-truth locations as follows: 1~km, 5~km, 10~km, 25~km, 50~km, 100~km, 200~km, 750~km and 2500~km.
Our evaluation focuses more on high accuracy range compared to the previous papers as we believe that fine-grained geolocalization is more important in practice.
A geolocational accuracy $a_r$ at a scale is given by the fraction of images in the test set localized within radius $r$ from ground-truths, which is given by
%
\begin{equation}
\label{eq:geo_acc}
a_r\equiv\frac{1}{M}\sum_{i=1}^{M}{u\left[ \mathrm{geodist}\left(l_\mathrm{gt}^i, l_\mathrm{pred}^i\right) < r \right]},
\end{equation}
where $M$ is the number of examples in the test set, $u[\cdot]$ is an indicator function and $\mathrm{geodist}(l_\mathrm{gt}^i, l_\mathrm{pred}^i)$ is the geolocational distance between the true image location $l_\mathrm{gt}^i$ and the predicted location $l_\mathrm{pred}^i$ of the $i$-th example.

\begin{table}[t]
\centering
\caption{Geolocational accuracies [\%] of models at different scales on Im2GPS3k.}
\label{tab:im2gps3k}
\scalebox{0.9}{
\begin{tabular}{
p{3.1cm}@{}|@{}C{1.cm}@{}@{}C{1.cm}@{}@{}C{1.cm}@{}@{}C{1.cm}@{}@{}C{1.cm}@{}@{}C{1.2cm}@{}@{}C{1.2cm}@{}@{}C{1.2cm}@{}@{}C{1.5cm}@{}
}
Models                                  & 1~km      & 5~km      & 10~km	    & 25~km     & 50~km     & 100~km    & 200~km    & 750~km    & 2500~km   \\ \hline\hline
ImageNetFeat                            & 3.0       & 5.5       & 6.4       & 6.9       & 7.7       & 9.0       & 10.8      & 18.5      & 37.5      \\ \hline
Deep-Ret~\cite{vo17revisiting}          & 3.7       & --        & --        & 19.4      & --        & --        & 26.9      & 38.9      & 55.9      \\
PlaNet~(reprod)~\cite{weyand16planet}   & 8.5       & 18.1      & 21.4      & 24.8      & 27.7      & 30.0      & 34.3      & 48.4      & \bf{64.6} \\ \hline
ClassSet~1                              & 8.4       & 18.3      & 21.7      & 24.7      & 27.4      & 29.8      & 34.1      & 47.9      & 64.5      \\
ClassSet~2                              & 8.0       & 17.6      & 20.6      & 23.8      & 26.2      & 29.2      & 32.7      & 46.6      & 63.9      \\
ClassSet~3                              & 8.8       & 18.9      & 22.4      & 25.7      & 27.9      & 29.8      & 33.5      & 47.8      & 64.1      \\
ClassSet~4                              & 8.7       & 18.5      & 21.4      & 24.6      & 26.8      & 29.6      & 33.0      & 47.6      & 64.4      \\
ClassSet~5                              & 8.8       & 18.7      & 21.7      & 24.7      & 27.3      & 29.3      & 32.9      & 47.1      & 64.5      \\ \hline
Average[1-2]                            & 8.2       & 18.0      & 21.1      & 24.2      & 26.8      & 29.5      & 33.4      & 47.3      & 64.2      \\
Average[1-5]                            & 8.5       & 18.4      & 21.5      & 24.7      & 27.1      & 29.5      & 33.2       & 47.4      & 64.3      \\ \hline
CPlaNet[1-2]                                 & 9.3       & 19.3      & 22.7      & 25.7      & 27.7      & 30.1      & 34.4      & 47.8      & 64.5      \\
CPlaNet[1-5]                                 & 9.9       & 20.2      & 23.3      & 26.3      & 28.5      & 30.4      & 34.5      & \bf{48.8} & \bf{64.6} \\ 
CPlaNet[1-5,PlaNet]                          & \bf{10.2} & \bf{20.8} & \bf{23.7} & \bf{26.5} & \bf{28.6} & \bf{30.6} & \bf{34.6} & 48.6      & \bf{64.6} \\
\hline
\end{tabular}}
\end{table}

\subsection{Results}

\subsubsection{Benefits of Combinatorial Partitioning}

Table~\ref{tab:im2gps3k} presents the geolocational accuracies of the proposed model on the Im2GPS3k dataset.
The proposed models outperform the baselines and the existing methods at almost all scales on this dataset.
ClassSet~1 through 5 in Table~\ref{tab:im2gps3k} are the models trained with the geoclass sets generated from the parameters presented in Table~\ref{tab:geoclass}.
Using the learned models as the base classifiers, we construct two variants of the proposed method---{CPlaNet}[1-2] using the first two base classifiers with manual parameter selection and CPlaNet[1-5] using all the base classifiers.

Table~\ref{tab:im2gps3k} presents that both options of our models outperform all the underlying classifiers at every scale.
Compared to na\"ive average of the underlying classifiers denoted by Average[1-5] and Average[1-2], {CPlaNet}[1-5] and {CPlaNet}[1-2] have $\sim$16 \% and $\sim$13 \% of accuracy gains at street level, respectively, compared to their counterparts.
We emphasize that {CPlaNet} achieves substantial improvements by a simple combination of the existing base classifiers and a generation of fine-grained partitions without extra training procedure.
The larger performance improvement in {CPlaNet}[1-5] compared to {CPlaNet}[1-2] makes sense as using more classifiers constructs more fine-grained geoclasses via combinatorial partitioning and increases prediction resolution.
Note that the number of distinct partitions formed by {CPlaNet}[1-2] is 46,294 while it is 107,593 in CPlaNet[1-5].

The combinatorial partitioning of the proposed model is not limited to geoclass sets from our generation methods, but is generally applicable to any geoclass sets.
Therefore, we construct an additional instance of the proposed method, {CPlaNet}[1-5,PlaNet], which also incorporates PlaNet~(reprod), reproduced version of PlaNet model~\cite{weyand16planet} with our training data, additionally.
{CPlaNet}[1-5,PlaNet] shows extra performance gains over {CPlaNet}[1-5] and achieves the state-of-the-art performance at all scales.
These experiments show that our combinatorial partitioning is a useful framework for image geolocalization through ensemble classification, where multiple classifiers with heterogeneous geoclass sets complement each other.

\begin{table}[t]
\centering
\caption{Geolocational accuracies [\%] on YFCC4k.}
\label{tab:yfcc4k}
\scalebox{0.9}{
\begin{tabular}{
p{3.1cm}@{}|@{}C{1.cm}@{}@{}C{1.cm}@{}@{}C{1.cm}@{}@{}C{1.cm}@{}@{}C{1.cm}@{}@{}C{1.2cm}@{}@{}C{1.2cm}@{}@{}C{1.2cm}@{}@{}C{1.5cm}@{}
}
Models	                                & 1~km	    & 5~km      & 10~km     & 25~km     & 50~km     & 100~km    & 200~km    & 750~km    & 2500~km	\\ \hline\hline
Deep-Ret~\cite{vo17revisiting}          & 2.3       & -         & -         & 5.7       & -         & -         & 11.0      & 23.5      & 42.0      \\
PlaNet (reprod)~\cite{weyand16planet}   & 5.6       & 10.1      & 12.2      & 14.3      &\bf{16.6}  &\bf{18.7}  & \bf{22.2} & \bf{36.4} & \bf{55.8} \\
CPlaNet[1-5]                                 & 7.3       & 11.7      & 13.1      &{14.7}     & 16.1      & 18.2      & 21.7      & 36.2      & 55.6      \\ 
CPlaNet[1-5,PlaNet]                          & \bf{7.9}  & \bf{12.1} & \bf{13.5} & \bf{14.8}  & 16.3     & 18.5      & 21.9      & \bf{36.4} & 55.5      \\
\hline
\end{tabular}
}
\end{table}

\begin{table}[t]
\centering
\caption{Geolocational accuracies [\%] on Im2GPS.}
\label{tab:im2gps}
\scalebox{0.9}{
\begin{tabular}{
@{}p{0.1cm}@{}@{}p{0.3cm}@{}|p{3.0cm}@{}|@{}C{1.cm}@{}@{}C{1.cm}@{}@{}C{1.cm}@{}@{}C{1.cm}@{}@{}C{1.cm}@{}@{}C{1.2cm}@{}@{}C{1.2cm}@{}@{}C{1.2cm}@{}@{}C{1.3cm}@{}
}
&&Models	                            & 1~km	        & 5~km      & 10~km     & 25~km     & 50~km     & 100~km    & 200~km    & 750~km    & 2500~km   \\ \hline\hline
&\multirow{4}{*}{\rotatebox[origin=c]{90}{\scriptsize{Retrieval}}}
 &Im2GPS~\cite{hays08im2gps}	        & - 	        & -         & -         & 12.0	    & -         & -         & 15.0	    & 23.0      & 47.0	    \\
&&Im2GPS~\cite{hays15large}	            & \phantom{0}2.5& 12.2      & 16.9      & 21.9      & 25.3      & 28.7      & 32.1      & 35.4      & 51.9      \\
&&Deep-Ret~\cite{vo17revisiting}	    & 12.2          & -         & -         & 33.3      & -         & -         & 44.3      & 57.4      & 71.3      \\
&&Deep-Ret+~\cite{vo17revisiting}	    & 14.4          & -         & -         & 33.3      & -         & -         & \bf{47.7} & 61.6      & 73.4      \\ \hline
&\multirow{6}{*}{\rotatebox[origin=c]{90}{\scriptsize{Classifier}}}
 &Deep-Cls~\cite{vo17revisiting}        & \phantom{0}6.8& -         & -         & 21.9      & -         & -         & 34.6      & 49.4      & 63.7  	\\
&&PlaNet~\cite{weyand16planet}          & \phantom{0}8.4& 19.0      & 21.5      & 24.5      & 27.8      & 30.4      & 37.6      & 53.6      & 71.3  	\\
&&PlaNet (reprod)~\cite{weyand16planet} & 11.0          & 23.6      & 26.6      & 31.2      & 35.4      & 30.5      & 37.6      & \bf{64.6} & \bf{81.9} \\ 
&&CPlaNet[1-2]	                            & 14.8          & 28.7      & 31.6      & 35.4      & 37.6      & 40.9      & 43.9      & 60.8      & 80.2      \\
&&CPlaNet[1-5]                               & 16.0          & \bf{29.1} & 33.3      & 36.7      & 39.7      & 42.2      & 46.4      & 62.4      & 78.5      \\
&&CPlaNet[1-5,PlaNet]                        & \bf{16.5}     & \bf{29.1} & \bf{33.8} & \bf{37.1} & \bf{40.5} & \bf{42.6} & 46.4      & 62.0      & 78.5      \\
\hline
\end{tabular}
}
\end{table}

We also present results on YFCC4k~\cite{vo17revisiting} dataset in Table~\ref{tab:yfcc4k}.
The overall tendency is similar to the one in Im2GPS3k.
Our full model outperforms Deep-Ret~\cite{vo17revisiting} consistently and significantly.
The proposed algorithm also shows substantially better performance compared to PlaNet (reprod) in the low threshold range while two methods have almost identical accuracy at coarse-level evaluation.

On the Im2GPS dataset, our model outperforms other classification-based approaches---Deep-Cls and PlaNet, which are single-classifier models with a different geoclass schema---significantly at every scale, as shown in Table~\ref{tab:im2gps}.
The performance of our models is also better than the retrieval-based models at most scales.
Moreover, our model, like other classification-based approaches, requires much less space than the retrieval-based models for inference.
Although Deep-Ret+ improves Deep-Ret by increasing the size of the database, it even worsens space and time complexity.
In contrast, the classification-based approaches including ours do not require extra space when we have more training images.

\begin{figure}[ht!]
\centering
\begin{subfigure}[b]{0.9\linewidth}
\centering
\includegraphics[width=1\linewidth] {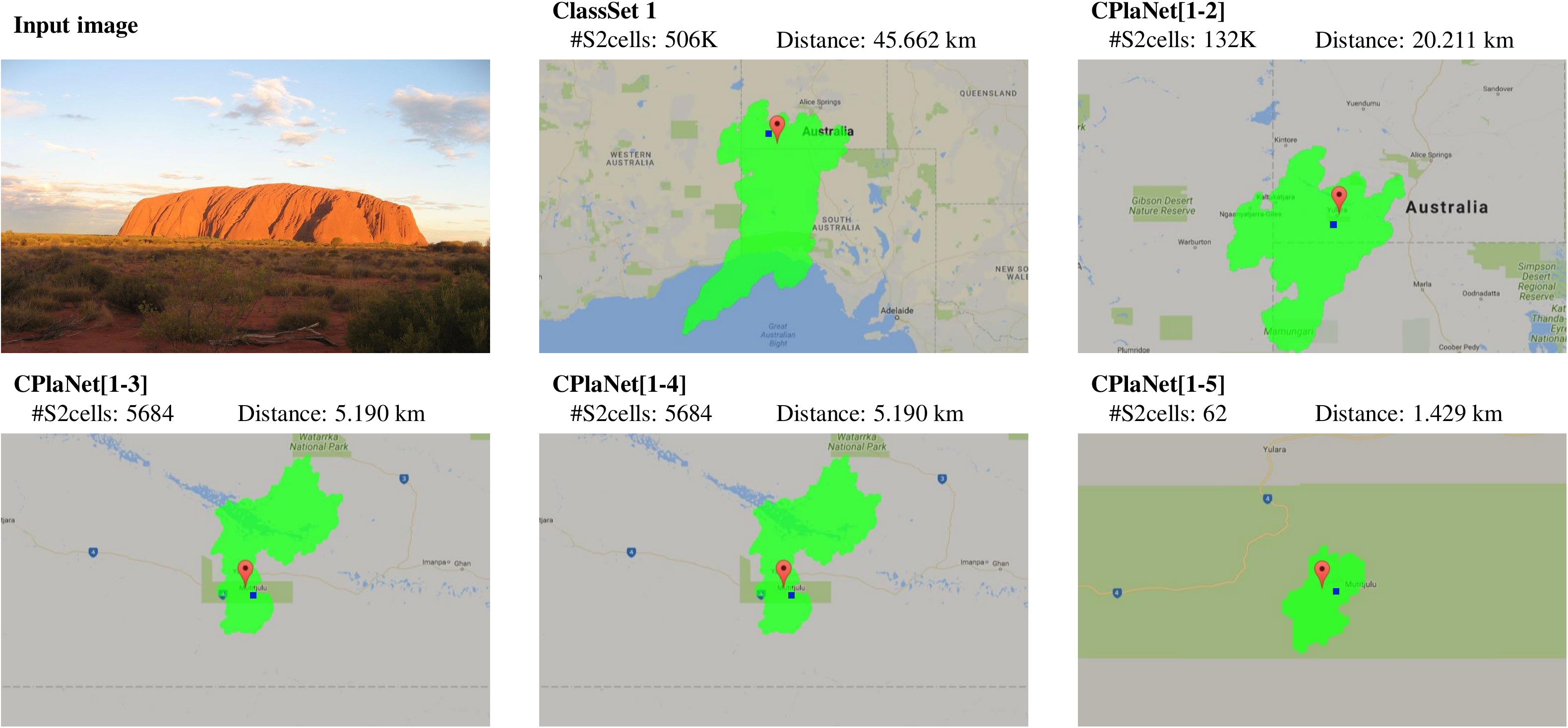}
\caption{Uluru in Australia}
\end{subfigure}

\begin{subfigure}[b]{0.9\linewidth}
\centering
\includegraphics[width=1\linewidth] {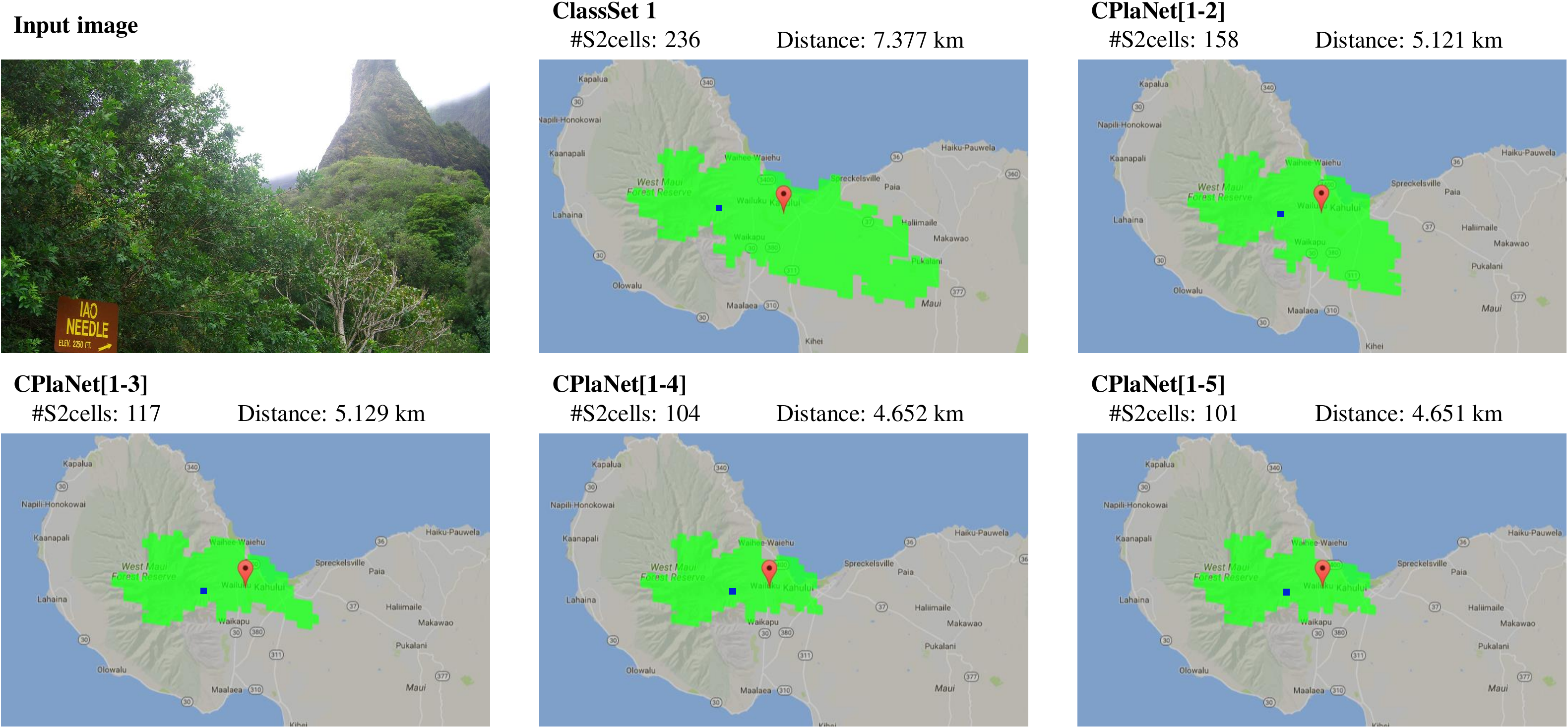}
\caption{Iao Needle in Hawaii}
\end{subfigure}
\caption{
Qualitative results of CPlaNet[1-5] on Im2GPS. 
Each map illustrates the progressive results of combinatorial partitioning by adding classifiers one by one.
S2 cells with the highest score and their centers are marked by green area and red pins respectively while the ground-truth location is denoted by the blue dots.
We also present the number of S2 cells in the highlighted region and distance between the ground-truth location and the center of the region in each map.
}
\label{fig:qualitative}
\end{figure}

Figure~\ref{fig:qualitative} presents qualitative results of {CPlaNet}[1-5] on Im2GPS.
It shows how the combinatorial partitioning process improves the geolocalization quality.
Given an input image, each map shows an intermediate prediction as we accumulate the scores on different geoclass sets one by one.
The region with the highest score is progressively sharded into a smaller region with fewer S2 cells, and the center of the region gradually approaches to the ground-truth location as we integrate more classifiers for inference.

\subsubsection{Computational Complexity}
Although {CPlaNet} achieves competitive performance through combinatorial partitioning, one may be concerned about potential increase of time complexity for its inference due to additional classification layers and overhead in combinatorial partitioning process.
However, it turns out that the extra computational cost is negligible since adding few more classification layers on top of the shared feature extractor does not increase inference time substantially and the required information for combinatorial partitioning is precomputed as described in Section~\ref{sub:inference}.
Specifically, when we use 5 classification branches with combinatorial partitioning, theoretical computational costs for multi-head classification and combinatorial partitioning are only 2\% and 0.004\% of that of feature extraction process.
In terms of space complexity, classification based methods definitely have great advantages over retrieval based ones, which need to maintain the entire image database.
Compared to a single-head classifier, our model with five base classifiers requires just four additional classification layers, which incurs moderate increase of memory usage.
%

\subsubsection{Importance of Visual Features}
For geoclass set generation, all the parameters of ClassSet~1 and 2 are set to the same values except for the relative importance of two factors for edge weight definition; edge weights for ClassSet~1 are determined by visual distances only whereas those for ClassSet~2 are based on geolocational distances between the cells without any visual information of images.
ClassSet~1 presents better accuracies at almost all scales as in Table~\ref{tab:im2gps3k}.
This result shows how important visual information of images is when defining geoclass sets.

Moreover, we build another model (ImageNetFeat) learned with the same geoclass set with ClassSet~1 but using a different feature extractor pretrained on ImageNet~\cite{Deng09ImageNet}.
The large margin between ImageNetFeat and ClassSet~1 indicates importance of feature representation methods, and implies unique characteristics of visual cues required for image geolocalization compared to image classification.

\begin{table}[t]
\centering
\caption{Comparisons between the models with and without normalization for combinatorial partitioning on Im2GPS3k.
Each number in parentheses denotes the geoclass set size, which varies largely to highlight the effect of normalization for this experiment.}
\label{tab:flickrgeo_normalization}
\scalebox{0.9}{
\begin{tabular}{
p{2.7cm}@{}|@{}C{1.cm}@{}@{}C{1.cm}@{}@{}C{1.cm}@{}@{}C{1.cm}@{}@{}C{1.cm}@{}@{}C{1.2cm}@{}@{}C{1.2cm}@{}@{}C{1.2cm}@{}@{}C{1.5cm}@{}
}
Models              & 1~km      & 5~km      & 10~km	    & 25~km     & 50~km     & 100~km    & 200~km    & 750~km    & 2500~km   \\ \hline\hline
ClassSet~1  (9969)  & 8.4       & 18.3      & 21.7      & 24.7      & 27.4      & 29.8      & 34.1      & 47.9      & 64.5      \\
ClassSet~2  (9969)  & 8.0       & 17.6      & 20.6      & 23.8      & 26.2      & 29.2      & 32.7      & 46.6      & 63.9      \\
ClassSet~3 (3416)   & 4.2       & 15.9      & 19.1      & 22.8      & 24.9      & 28.0      & 31.4      & 46.1      & 63.5      \\
ClassSet~4 (1444)   & 1.8       & 9.5       & 13.2      & 16.8      & 21.2      & 24.5      & 29.5      & 44.4      & 61.8      \\
ClassSet~5 (10600)  & 8.2       & 19.1      & 22.3      & 25.2      & 27.3      & 29.9      & 33.6      & 47.3      & \bf{65.5} \\ \hline
SimpleSum           & 9.7       & 19.4      & 23.1      & 26.6      & 28.1      & 30.6      & 33.8      & 47.7      & 64.0      \\
NormalizedSum       & \bf{9.8}  & \bf{19.8} & \bf{23.6} & \bf{26.8} & \bf{28.8} & \bf{31.1} & \bf{34.9} & \bf{48.3} & 65.0      \\ \hline
\end{tabular}
}
\end{table}

\subsubsection{Balancing Classifiers}
We normalize the scores assigned to individual S2 cells as discussed in Section~\ref{sub:inference}, which is denoted by NormalizedSum, to address the artifact that sums of all S2 cell scores are substantially different across classifiers.
To highlight the contribution of NormalizedSum, we conduct an additional experiment with classsets that have large variations in number of classes.
Table~\ref{tab:flickrgeo_normalization} presents that NormalizedSum clearly outperforms the combinatorial partitioning without normalization (SimpleSum) while SimpleSum still illustrates competitive accuracy compared to the base classifiers.

%% file: sections/conclusion.tex

\section{Conclusion}
\label{sec:conclusion}

We proposed a novel classification-based approach for image geolocalization, {referred to as CPlaNet}.
Our model obtains the final geolocation of an image using a large number of fine-grained regions given by combinatorial partitioning of multiple classifiers.
We also introduced an inference procedure appropriate for classification-based image geolocalization.
The proposed technique improves image geolocalization accuracy with respect to other methods in multiple benchmark datasets especially at fine scales, and also outperforms the individual coarse-grained classifiers.